\documentclass[letterpaper, 10 pt, conference]{ieeeconf}  

\IEEEoverridecommandlockouts                              

\usepackage{bm}  
\usepackage{graphics} 
\usepackage{epsfig} 
\usepackage{times} 
\usepackage{amsmath} 
\usepackage{amssymb}  
\usepackage{amssymb}
\usepackage{color}
\usepackage{booktabs}
\usepackage{caption}
\usepackage{subcaption}
\usepackage{float}
\usepackage[ruled,linesnumbered]{algorithm2e}
\usepackage{algorithmic}

\usepackage[bookmarks=false]{hyperref}
\usepackage[noadjust]{cite}

\title{\LARGE \bf
Harnessing with Twisting: Single-Arm Deformable Linear Object Manipulation for  Industrial Harnessing Task
}

\author{Xiang Zhang$^{1}$, Hsien-Chung Lin$^{2}$, Yu Zhao$^{2}$, and Masayoshi Tomizuka$^{1}$
\thanks{$^{1}$Mechanical Systems Control Lab, UC Berkeley, USA.
        {\tt\small \{xiang\_zhang\_98, tomizuka\}@berkeley.edu}}%
\thanks{$^{2}$ FANUC Advanced Research Laboratory, FANUC America Corporation, USA.
        {\tt\small \{john.lin,Yu.Zhao\}@fanucamerica.com}}%
}

\begin{document}

\maketitle
\thispagestyle{empty}
\pagestyle{empty}

\begin{abstract}
Wire-harnessing tasks pose great challenges to be automated by the robot due to the complex dynamics and unpredictable behavior of the deformable wire. Traditional methods, often reliant on dual-robot arms or tactile sensing, face limitations in adaptability, cost, and scalability. This paper introduces a novel single-robot wire-harnessing pipeline that leverages a robot's twisting motion to generate necessary wire tension for precise insertion into clamps, using only one robot arm with an integrated force/torque (F/T) sensor. Benefiting from this design, the single robot arm can efficiently apply tension for wire routing and insertion into clamps in a narrow space. Our approach is structured around four principal components: a Model Predictive Control (MPC) based on the Koopman operator for tension tracking and wire following, a motion planner for sequencing harnessing waypoints, a suite of insertion primitives for clamp engagement, and a fix-point switching mechanism for wire constraint updating. Evaluated on an industrial-level wire harnessing task, our method demonstrated superior performance and reliability over conventional approaches, efficiently handling both single and multiple wire configurations with high success rates. 
\end{abstract}

\section{Introduction}

Recent advancements in robotics have seen significant progress in handling rigid objects \cite{zhang2022learning, zhang2023learning, zhou2022learning}, but extending these advanced dexterous skills to manipulate deformable objects like cables, wires, and clothing remains a challenge. This challenge arises primarily due to the complex behavior of deformable objects, which have a high-dimensional state space and nonlinear dynamics that complicate state estimation, path planning, and predictive modeling, making automation in real-world scenarios difficult. The complexity increases when these objects interact with their environment, as seen in tasks like wire harnessing \cite{galassi2021robotic, suberkrub2022feel, chen2024real,wilson2023cable,jin2022robotic, yu2024hand}.

Fig.~\ref{fig:taskboard} illustrates a typical wire harnessing setup, where a robot must shape a deformable linear object (DLO) - in this case, a wire - into a specific form and secure it with clamps. Unlike other deformable object manipulation tasks such as shape tracking \cite{wang2022offline, yu2023generalizable} or cloth folding \cite{avigal2022speedfolding}, wire harnessing demands accurate state estimation to determine the wire's shape and requires precise control over the force applied to tension the wire for insertion into clamps. Previous solutions have utilized dual-robot arms to create tension \cite{suberkrub2022feel,chen2024real, jin2022robotic} or tactile sensors \cite{galassi2021robotic,wilson2023cable} to gauge gripping and shearing forces, but these approaches have limitations. Dual-arm setups are costly and pose motion planning challenges, while tactile sensors can degrade over time \cite{donlon2018gelslim}, losing effectiveness. Moreover, most prior research has focused on simplified scenarios with either loose clamps \cite{wilson2023cable} that reduce the harnessing to a placing task or large gaps between clamps that make it easier for the robot to generate tension. Thus, the previous settings may not accurately reflect the complexities of real-world applications, such as the NIST board assembly challenge \cite{NIST}.


Our approach aims to address this complex task using a single robot arm equipped with force sensing. We leverage a simple yet effective strategy involving stretching and twisting motions. Stretching straightens the wire, simplifying state estimation and preparing it for insertion into a clamp, while twisting twines the wire to the robot gripper and adjusts the tension by changing the twisting angle. This combination allows a single arm to precisely position and tension the wire for insertion. Our proposed framework comprises four main components: 1) a Model Predictive Control (MPC) module that uses a Koopman-operator based dynamics function for motion planning and tension prediction, 2) a waypoint planning module for mapping out the path and avoiding collisions, 3) insertion primitives designed for securing the wire in clamps without losing tension, and 4) a fix-point switching mechanism for wire constraint updating. We evaluated our framework in the NIST board challenge, a task that represents an industrial-level wire harnessing challenge. Our system demonstrated a high success rate in real-world wire-following trajectories, showcasing its potential to effectively tackle the demanding wire-harnessing task while managing tension forces.
\section{Related Work}
\subsection{Challenges of Deformable Manipulation}
Manipulation of deformable objects has recently garnered significant interest in robotics research, with state estimation and dynamics modeling emerging as key challenges in automating such tasks.

\textbf{State estimation} is crucial for subsequent manipulation tasks. Traditional methods \cite{tang2018framework, tang2022track, jin2020real} often rely on color filtering to distinguish the object from its background, followed by non-rigid registration methods to derive a keypoint representation for planning and manipulation. This approach, however, depends heavily on manual tuning of filter parameters and is not robust to changes in task settings. Recent advancements have seen learning-based methods excel in identifying states of deformable objects. For instance, Jin et al. \cite{jin2022robotic} introduced a deep neural network (DNN) that leverages real-world data to detect deformable linear objects (DLOs) in an environment. Similarly, Yan et al. \cite{yan2020self} employed a coarse-to-fine strategy using VGG feature maps to locate rope segments, enhancing resolution progressively. Furthermore, some end-to-end models \cite{nair2017combining} bypass the need for explicit state estimation by directly using visual inputs to predict manipulative actions or the dynamics of deformable objects. In our research, we simplify the state estimation challenge by straightening the wire, thereby facilitating easier handling and application of tension for the wire harnessing task.

\textbf{Dynamics modeling} of deformable objects is particularly difficult due to their infinite degrees of freedom. A common simplification involves representing these objects with a finite set of particles or connected rigid bodies \cite{tang2022track, wang2022offline}, though this often fails to accurately capture force interactions and can lead to non-intuitive model parameters \cite{yin2021modeling}. Alternatively, learning-based methods have been developed to approximate the dynamics of deformable objects from data \cite{nair2017combining, yan2020self, wang2022offline}. These models, depending on their state representations, can predict future states of the object, whether as key points or entire images, based on the robot's actions. The choice of network, whether MLP, LSTM, or, as Wang et al. \cite{wang2022offline} have shown, graph neural networks (GNNs), significantly impacts performance. GNNs, in particular, are effective due to their structure, which inherently captures the dynamics of deformable objects. In our work, we simplify the system by applying tension to the wire, using the Koopman operator for dynamic modeling, which accelerates training and inference, enabling real-time optimal control.
\begin{figure}[t!]
    \centering
    \includegraphics[width=0.45\textwidth]{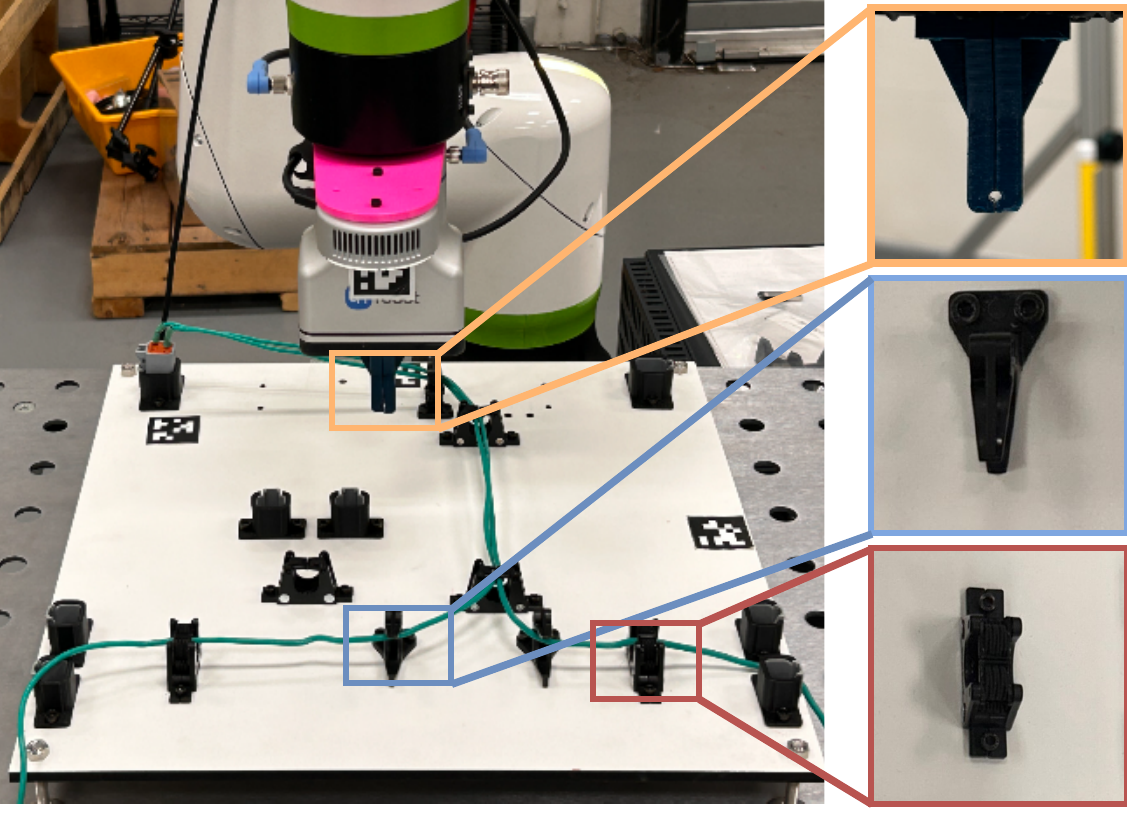}
    \caption{NIST task board setup for wire harnessing task. We use a 3D-printed finger with grooves (orange box) to harness the wire into ``C"-shaped (blue box) and ``U"-shaped clamps (red box).}
    \label{fig:taskboard}
\end{figure}
\subsection{Wire Harnessing Task}
Wire harnessing presents a complex challenge in deformable object manipulation, requiring precise contact with the environment and deformation into specific shapes. The primary challenges are twofold: guiding the wire through a series of clamps and securing it within these clamps. Traditional approaches have explored dual-arm setups \cite{suberkrub2022feel,chen2024real, jin2022robotic} or tactile sensors \cite{galassi2021robotic, wilson2023cable}. In a dual-arm configuration, one arm holds the wire steady while the other applies tension for routing and insertion. Tactile setups rely on sensors to detect the wire's pose and tension for guidance and insertion. However, these methods are either complex and costly or suffer from durability issues with tactile sensors. Our approach uses a single robot arm equipped with force-torque (FT) sensing, utilizing stretching and twisting motions to generate tension, simplifying state estimation and aiding in both routing and insertion tasks. The closest work to this paper is \cite{suberkrub2022feel}, whose authors adopted a dual-arm setup with FT sensing and occasionally uses single arm to stretch the wire for insertion. However, our setup is more challenging since the clamps are much smaller and tighter than their settings which requires the robot to generate enough tension in a limited space to finish the insertion. In the experiments, we demonstrate our approach can achieve faster force tracking performance with twisting motion and successes in this challenging real-world harnessing task.

\section{Problem Formulation}
\subsection{Wire Harnessing Task Set-up}
\begin{figure}
    \centering
    \includegraphics[width=0.48\textwidth]{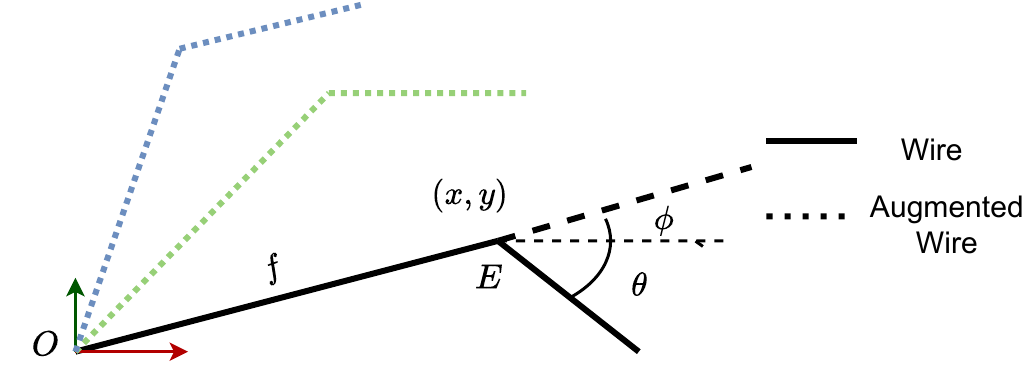}
    \caption{State space of the wire harnessing task. $O$ represents the fixed wire origin. $E$ is the robot end-effector (gripper) to bend and twist the wire. The state space consists of $(x,y,\theta,f)$ denotes the robot position and rotation relative to $O$ and the tension force. $\phi$ is the twisting angle between the wire and gripper. }
    \label{fig:state}
\end{figure}
The NIST Assembly Task Board 4 \cite{NIST} serves as our testbed for the wire harnessing task, incorporating both connector assembly and wire harnessing tasks. Since our focus is not on the connector assembly, we proceed under the assumption that the connector is pre-attached to the board, and one end of the wire is already fixed to this connector. The robot's objective is then to route the wire into a desired configuration by securing it within a series of clamps. The board features two clamp types: ``C"-shaped clamps act as pivot points for wire routing, and ``U"-shaped clamps secure the wire post-insertion. With the 3D geometry of the NIST task board known, we assume that each clamp's relative position is predetermined, and the board's location is identifiable via an AR tag, simplifying the vision-based clamp localization process. For environments where clamp detection is necessary, the segmentation and pose estimation method mentioned in \cite{wilson2023cable} could be applied. Fig.~\ref{fig:taskboard} illustrates the task board and an example of the targeted wire configuration.


\subsection{Robot Setup for Wire Harnessing}
\textbf{Robot Manipulator:} Diverging from the dual-arm or tactile sensor-based approaches of previous studies, our method requires only a single robot arm equipped with force sensing capabilities. We employ a FANUC CRX-10\textit{i}A collaborative robot for our experiments. Equipped with joint torque sensors, the robot can directly estimate the tension force during harnessing using the relationship $F = J^T \tau$, where $\tau \in \mathbb{R}^6$ represents the joint torque, $J \in \mathbb{R}^{6 \times 6}$ denotes the Jacobian matrix, and $F \in \mathbb{R}^6$ is the external wrench. For other robots without this capability, an external F/T sensor on the robot wrist can be utilized to directly measure the tension force $F$.

A RealSense L515 camera attached to the end-effector aids in acquiring the task board's pose and generating the task board's point cloud for initial wire grasping pose estimation.

\begin{figure*}[t!]
    \centering
    \includegraphics[width=0.9\textwidth]{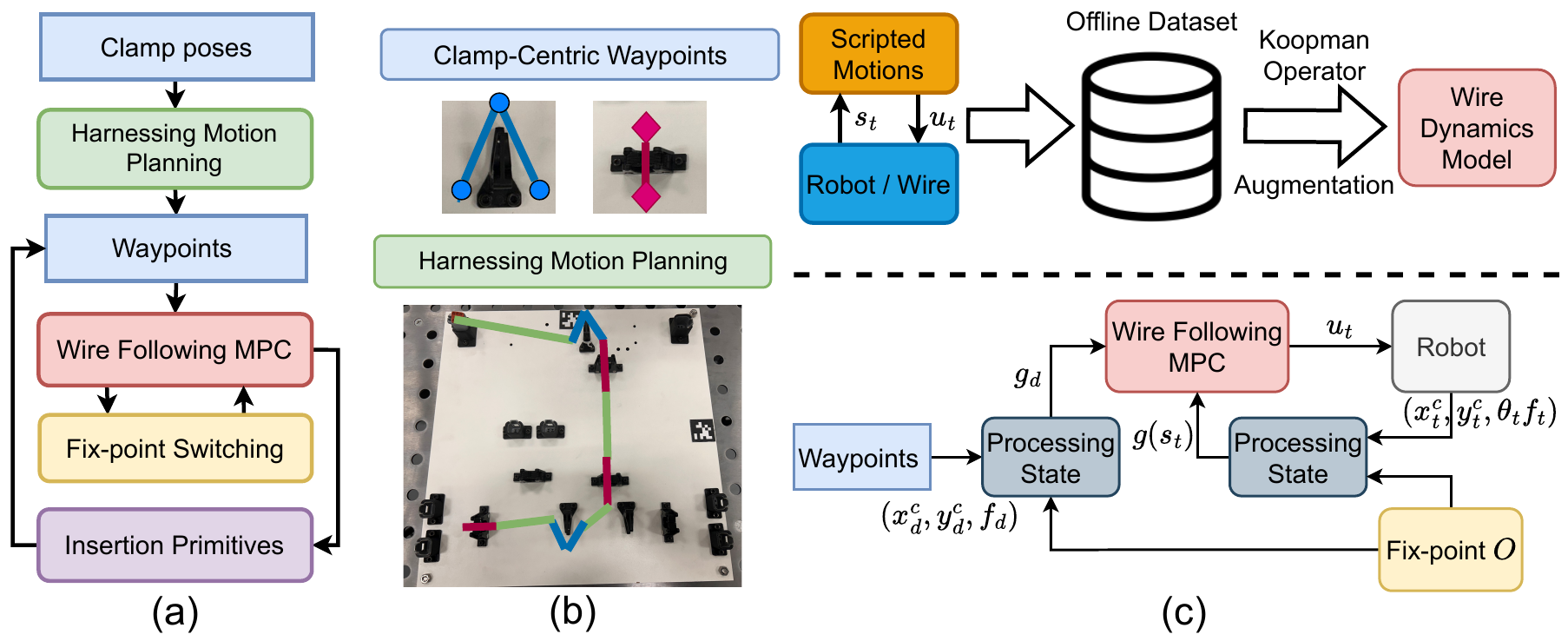}
    \caption{The overview of our proposed approach: a) the pipeline for wire-harnessing, b) harnessing motion planer that sequences and merges a set of clamp-centric waypoints for each of the clamps, c) Top: We use pre-scripted motions to collect real-world wire following data, then augment by 10 times for Koopman operator fitting. Bottom: We first process the waypoints and robot state in the Cartesian space to the fix-point frame and use Koopman lift function to obtain the wire state. Then we utilize MPC to infer the optimal control command.}
    \label{fig:framework}
\end{figure*}

\textbf{Gripper Design:} The standard parallel gripper, while commonly used, proves unsuitable for tasks requiring wire stretching or following due to the unpredictability of in-hand wire pose and the risk of wire slippage. As shown in Fig.~\ref{fig:taskboard}, to overcome this, we redesigned the gripper to better suit our needs, incorporating a thin finger design with grooves that allow the wire to pass through, thereby securing it during harnessing and enhancing task robustness. The robot manipulates wire tension by actively twisting it, applying rotational motion to increase friction.


\subsection{State and Action Spaces Design}

\textbf{State space:} The wire's state in the harnessing task can exhibit hybrid dynamics: either slack, exhibiting no tension and assuming random shapes, or stretched, forming a straight line under tension. Given that one wire end is always fixed (either to the connector or by clamps), and assuming pre-grasp by the robot in a planner stretching scenario, we consider a simplified 2D case as shown in Fig.~\ref{fig:state}, where the origin $O$ is the fix-point and $E$ is the robot end-effector, a gripper, twisting the wire (solid line). We define the wire state as $s = (x, y, \theta, f)^T$, where $x, y, \theta$ represent the robot gripper's relative position to the fix-point $O$ and $Z$ axis rotation, respectively, and $f = ||F||_2$ indicates the tension force's 2-norm. This 2D case simplification omits $Z$ direction motion, reducing system complexity. Additionally, since the tension force direction aligns with the robot's pose, only the force magnitude is considered. Since the wire will be twisted and stretched throughout the harnessing process, this setup directly utilizes the robot gripper's position to monitor the wire's state.

\textbf{Action space:} The robot's actions in wire harnessing combine stretching and twisting motions. Specifically, the control command $u$ includes translational incremental motion $(\Delta x, \Delta y)^T$ for stretching and rotational incremental motion $\Delta \theta$ for twisting. Each command will be executed for $0.5~s$ by the robot. Throughout the task, the robot adjusts its actions to stretch and twist the wire, achieving the necessary tension for task completion.

\subsection{Basics of Koopman Operator}
Consider a discrete nonlinear dynamics system $s_{t+1}=F(s_t)$, where $s_t \in \mathbb{R}^n$, the Koopman operator: $\mathcal{K} : \mathcal{F} \rightarrow \mathcal{F}$ is an infinite linear transformation that evolves every observables $g:\mathbb{R}^n \rightarrow \mathbb{R}$ belonging to $\mathcal{F}$ \cite{mauroy2020koopman}:
\begin{equation}
(\mathcal{K}g) (s_t) = g(F(s_t)) = g(s_{t+1})
\end{equation}

In practice, we use a finite approximation of the infinite linear transformation with Koopman matrix $K \in \mathbb{R}^{m\times m}$ and use a lift function $g(x_t):\mathbb{R}^n \rightarrow \mathbb{R}^m$ to represent a set of base observation functions $(g_1,\dots, g_m)$. While the original Koopman operator does not consider the control term $u_t \in \mathbb{R}^l$, previous works \cite{bruder2019modeling, brunton2016koopman, li2019learning}have shown it can be linearly combined with the Koopman observation space resulting in the form of:
\begin{equation}
    g(s_{t+1}) = K g(s_t) + L u_t
\end{equation}
where $L \in \mathbb{R}^{m \times l}$ is the control matrix.

\section{Proposed Approach} \label{sec:proposed}

Our key idea employs stretching and twisting motions to generate the tension required for wire harnessing with a single arm. This approach consists of four main components: 1) Wire following with force tracking control, which plans stretching and twisting motions to guide the wire to the intended pose while maintaining tension; 2) Harnessing motion planning, which sequences the waypoints to complete the harnessing task; 3) Clamp insertion primitives, which use the wire's tension for precise alignment with ``U"-shaped clamps; and 4) a fix-point switching mechanism that updates the current wire constraints. The whole pipeline is depicted in Fig.~\ref{fig:framework}(a).

\subsection{Wire Following with Force Tracking Control}
Achieving the right wire tension is crucial, as excessive force can damage the wire or connector, while a slack wire with insufficient tension makes clamp insertion challenging. The complexity lies in the wire's nonlinear dynamics and the difficulty of determining optimal control commands. We address this by leveraging the Koopman operator for a linear dynamic model in the lift space, using real-world wire following trajectories. An MPC module then calculates the optimal robot movements for tension force tracking and waypoint navigation.


\textbf{Koopman dynamics fitting:}
To fit the nonlinear wire dynamics, we define the lifting function $g(s_t)$ as follows:
\begin{equation}
    g(s_t) = (s_t, \phi_t, z(s_t))^T = (x_t, y_t, \theta_t, f_t, \phi_t, z(s_t))^T
\end{equation}
, where $\phi_t$ represents the twisting angle between wire and gripper at time $t$ as shown in Fig.~\ref{fig:state}, and $z(s_t)$ includes the 2-order polynomial of $(s_t, \phi_t)$. We chose this setup for its simplicity and adequacy in representing wire dynamics. We offline collected 40 real-world wire-following trajectories $[\tau_i = (s_0,u_0,\dots, s_T,u_T), i=1,\dots,40]$ as the dataset using scripted twisting and stretching motion with random initial states. To fit the model, our objective is to minimize one-step prediction loss:
\begin{equation}
    L = \sum_{i=1}^{N-1} \left\Vert g(s_{i+1}) - \begin{bmatrix}
           K, L
         \end{bmatrix}\begin{bmatrix}
           g(s_i) \\
           u_i
         \end{bmatrix}\right\Vert^2_2
\end{equation}
To simplify, we denote $\begin{bmatrix}g(s_i) \\u_i\end{bmatrix}$ as $g'(s_i,u_i)$. Then, the Koopman and control matrix can be obtained by using the closed-form solution:
\begin{align}
   &\begin{bmatrix}
           K, L
         \end{bmatrix} = PG^\dagger \\
    &P=\frac{1}{N-1}\sum_{i=1}^{N-1} g(s_{i+1})g'(s_i)^T\\ 
    &G=\frac{1}{N-1}\sum_{i=1}^{N-1} g'(s_{i})g'(s_i)^T
\end{align}
where $\dagger$ is the Moore–Penrose pseudoinverse.


\textbf{Geometric inspired data augmentation:}
Since we only have a limited number of training data, we also augmented them with geometric heuristics to improve the model performance. As shown in Fig.~\ref{fig:state}, our intuition is that the tension force will be identical if we rotate the stretching trajectory by an angle $\psi$. In practice, we augment each real-world trajectory by $10$ times with a rotation angle from $[-\pi,\pi]$ with an interval of $\frac{1}{5}\pi$. 

\textbf{MPC setup:} We utilize MPC to infer the optimal control command by solving the following optimization:

\begin{equation}
\begin{aligned}
\min_{u_i,\dots, u_{i+H}} \quad & \sum_{t=i}^H (g(s_t)-g_d)^T Q (g(s_t)-g_d) + u_t^T R u_t\\
\textrm{s.t.} \quad & g(s_{t+1}) = K g(s_t) + L u_t\\
  & b_l \leqslant A g(s_{t+1})\leqslant b_u \\
  & c_l \leqslant u_t\leqslant c_u, \forall t \in [i, i+H]
\end{aligned}
\end{equation}
where $g_d = (x_d, y_d, 0, f_d, 0, \dots, 0)^T$ represents the desired waypoint $(x_d, y_d)$ and desired force $f_d$. $Q = diag(10,10,0,1,0,\dots,0)$ focuses on penalizing only position and force tracking errors and ignores the rotation term, $R = diag(0.1,0.1,0.1)$ regulates control effort, $A = [\mathbb{I}^{4}, 0]$ is the selection matrix to extract the original state and $[b_l, b_u],[c_l,c_u]$ are lower and higher bounds for the state and control, respectively. We keep solving this problem for each step and only execute $u_i$ for each time. 

One thing to note is that the origin of the state is located at the fix-point but the way-points and the robot poses are planned or measured in the world frame. Therefore, at each control loop,  we process the Cartesian space waypoint $(x_d^c, y_d^c,f_d)$, and robot state $(x_t^c, y_t^c, \theta_t, f_t)$ to the fix-point frame and then project them to the lift space to run the MPC.

\subsection{Harnessing Motion Planning}

Given a sequence of clamps for wire harnessing, we propose a clamp-centric waypoint planning algorithm. As shown in Fig.~\ref{fig:framework}(b), for ``U"-shape clamps, there are two waypoints for the robot to pull the wire on the top of the clamp and secure it with the insertion primitive. ``C"-shape clamps have two side-waypoints and one tip-waypoint to route the wire around the clamp. The overall planning consists of two steps. In the first step, starting from the initial fix-point, we add the waypoints of the next clamp based on their distance to the most current waypoint. For ``C"-shape clamps, the tip-waypoint must be in the middle of two side-waypoints. After obtaining a set of waypoints to follow, step two combines the waypoints that are within $20~mm$ distance to their mean to improve the efficiency of the planned path. The desired force $f_d$ is assigned to $7~N$ for ``C"-shaped clamp waypoints for routing and $10~N$ for ``U"-shaped clamp waypoints since the insertion requires a greater tension.

\subsection{Clamp Insertion Primitives}
\begin{figure}
    \centering
    \includegraphics[width=0.48\textwidth]{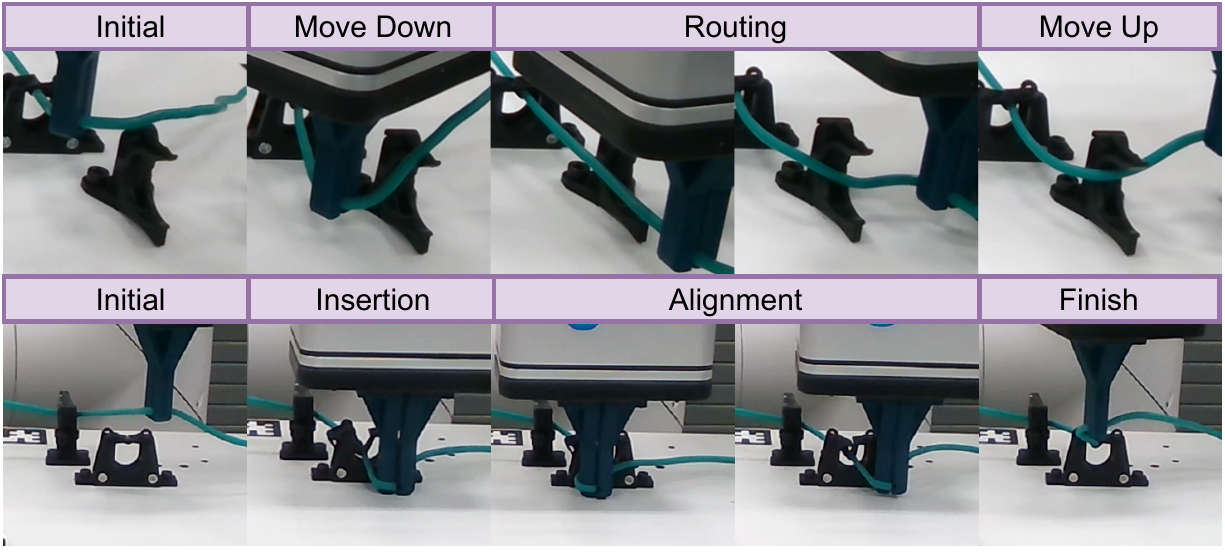}
    \caption{Insertion primitives for ``C"-shaped clamps (top) and ``U"-shaped clamps}
    \label{fig:primitives}
\end{figure}
When the robot stretches the wire to the desired waypoint, we utilize designed insertion primitives to align it to the clamps. Fig.~\ref{fig:primitives} depicts the snapshots of the primitive and their details are explained below:

\textbf{``C"-shaped clamps:} For the first side-waypoint, the robot moves down by $30~mm$ to lower its position for routing. After reaching the second side-waypoint, the robot moves up again for $30~mm$  to avoid collision with the board and finishes the routing.

\textbf{``U"-shaped clamps:} After reaching the second side-waypoint, the robot first goes down $30~mm$ for insertion while keeping twisting to maintain the wire tension. Then the robot moves along the edge of the clamp in two directions for $20~mm$ to further secure the wire. Finally, the robot moves up $30~mm$ to finish the insertion.

\subsection{Fix-point Switching}

\begin{figure}
    \centering
    \includegraphics[width=0.4\textwidth]{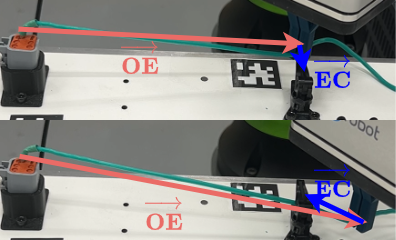}
    \caption{Fix-point switching mechanism.}
    \label{fig:fix_point}
\end{figure}

\begin{figure}[http]
    \centering
    \includegraphics[width=0.48\textwidth]{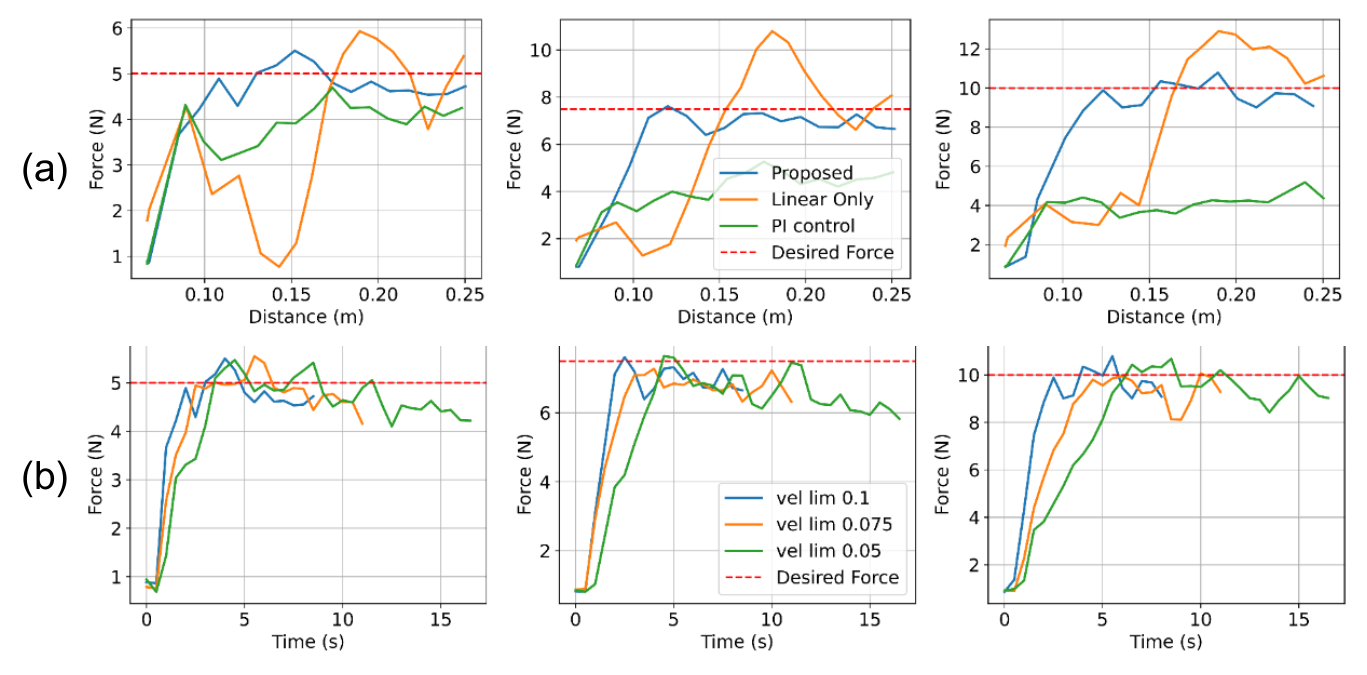}
    \caption{Tension tracking comparison of a) Proposed approach against baselines and b) Effect of different tracking velocity limits}
    \label{fig:force_tracking}
\end{figure}
Upon successful wire alignment, we update the fix-point, which is the origin of the state. Fig.~\ref{fig:fix_point} depicts the routing process for a ``C"-shaped clamp, where $O$ is the fix-point, $E$ is the gripper, and $C$ stands for the clamp. For ``C"-shaped clamps, the wire's binary position relative to the clamp is determined by $\overrightarrow{OE} \times \overrightarrow{EC} \cdot (0,0,1)^T$, which indicates the wire is on the left or right-hand side of the clamp. If $||OE|| > ||OC||$ which means the wire is stretched long enough to route on the clamp and the binary wire position changes the sign, we then switch the fix-point $O$ to the the clamp $E$. For ``U"-shaped clamps, we change the fix-point when the insertion primitive is executed.
\begin{figure*}[!htb]
    \centering
    \includegraphics[width=\textwidth]{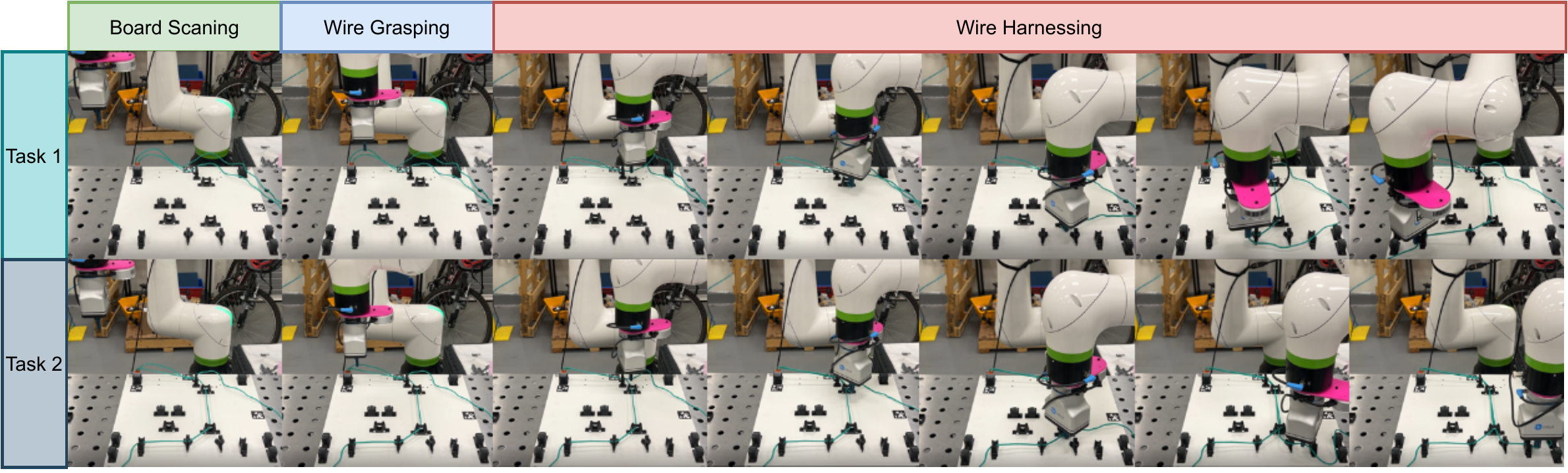}
    \caption{Snapshots of the multi-wire harnessing task, where the robot needs to harness two wires in a sequence. The robot first captures the workspace point cloud to identify and grasp the wire. Then, the proposed harnessing motion is applied to align the wire to the fixtures. Task 1 and 2 represent two different desired wire shapes and are executed sequentially.}
    \label{fig:snapshots}
\end{figure*}

\section{Experiments}

In this section, we evaluate the wire-harnessing pipeline proposed in Section~\ref{sec:proposed} on the NIST assembly task board. To assess the effectiveness of the twisting motion in generating cable tension and the necessity of a Koopman operator-based dynamics model, we compare our method against two baselines: \textbf{No-twisting}, which adjusts stretching velocity using a PI controller to generate wire tension, and \textbf{Linear dynamics}, which employs a linear dynamics model $s_{t+1} = A s_{t} + Bu_t$ with the same MPC setup for optimal robot motion determination. Specifically, we aim to address three questions related to wire-harnessing performance: 1) Can our Koopman-based MPC generate the desired wire tension? 2) Is the proposed pipeline capable of completing the complex wire-harnessing task? and 3) Can our method be extended to handle multiple wires?

\subsection{Tension Tracking Performance} \label{subsec:tension_tracking_exp}

To address the first question, we examined our Koopman-based MPC's tension-tracking capability. In our tests, we presuppose the robot is grasping the wire, and then stretching it by $250~mm$ in the $Y$ direction to sustain tension forces of $[5,7.5,10]~N$. Initially, we set the twisting angle $\phi$ to $0^\circ$ and capped the robot's maximum velocity at $0.1 m/s$ and $6^\circ/s$ for translational and rotational movements, respectively. The force-tracking outcomes are showcased in Fig.~\ref{fig:force_tracking}(a). Our findings indicate that the Koopman-based MPC proficiently maintains the desired wire tension, unlike the No-twisting benchmark, which caps at a $4~N$ tension force since the robot velocity is bounded by $0.1 m/s$ and $4~N$ is the maximum frictional force that could be applied without the twisting motion. Therefore, it fails to track a greater tension force. The linear dynamics benchmark, on the other hand, struggles with stable tension tracking due to its inability to accurately capture the wire's complex tension dynamics, underscoring the superiority of our proposed method.

Further investigation into harnessing velocity effects shows that varying maximum robot speeds yield comparable tension tracking performances, with a velocity of $0.1 m/s$ facilitating the quickest task completion as shown in Fig.~\ref{fig:force_tracking}(b). Hence, this velocity limit was adopted for all subsequent wire-harnessing tasks.

\subsection{Single-Wire Harnessing Task} \label{subsec:single_exp}
Next, we evaluated our complete proposed pipeline on the wire harnessing task. As illustrated in Fig.~\ref{fig:snapshots}, the robot first identifies the task board's pose using an April Tag. Subsequently, the motion planning module designates a series of waypoints for the robot to follow. To infer the grasping pose, the robot captures a multi-view point cloud of the wire near the initial fixed point using its in-hand camera and then selects the highest point as the grasp location. This straightforward yet effective heuristic ensures reliable wire grasping, applicable even in scenarios involving multiple wires. After grasping, the robot proceeds along the planned waypoints, maintaining wire tension via the Koopman-based MPC, updating the fix-point as necessary, and employing insertion primitives to anchor the wire within the clamps.

\begin{table}[h!]
  \centering
  
  \begin{tabular}{ccc|cc}
    \toprule
             & T1 Succ.Rate & Failures & T2 Succ.Rate & Failures\\
    \hline
    Proposed &10/10 & N/A & 10/10 & N/A\\
    No-Twist &0/10 & B(10) & 0/10 & B(10)\\
    Linear & 5/10& A(2)B(2)D(1) & 7/10 & B(2)C(1)\\
    \bottomrule
  \end{tabular}
  \caption{\textbf{Single wire-harnessing success rates:} Our proposed approach and baselines were tested across two harnessing tasks in 5 trials each. Failure modes include: A (wire tangled with clamps), B (failed to insert the wire into clamps), C (wire was pulled out of the connector), and D (wire slipped from the gripper due to excessive force).}
  \label{table:tasks}
\end{table}
We designed two harnessing tasks that navigate the wire from the same starting point to either a left or right route, depicted in Fig.\ref{fig:taskboard}. Table.~\ref{table:tasks} presents the overall performance. Remarkably, our proposed method successfully completed all 10 trials for both tasks, significantly outperforming the baselines. The No-twisting approach, as detailed in Section~\ref{subsec:tension_tracking_exp}, could not generate sufficient tension using stretching alone, failing to robustly insert the wire into the clamps. Likewise, the Linear model struggled with stable force tracking, occasionally applying excessive tension or completely losing tension, making the wire difficult to insert into the clamps or slip out of the gripper. These outcomes further affirm the Koopman operator's utility in capturing the intricate dynamics of wire tension.


\subsection{Multi Wire Harnessing Task}
\begin{table}[h!]
  \centering
  \begin{tabular}{ccc|cc}
    \toprule
             & T1 Succ.Rate & Failures & T2 Succ.Rate & Failures\\
    \hline
    Proposed &9/10 & A(1) & 8/9 & A(1)\\
    No-Twist &0/10 & B(10) & N/A & N/A\\
    Linear & 8/10& B(1)D(1) & 6/8 & A(1)D(1)\\
    \bottomrule
  \end{tabular}
  \caption{\textbf{Multi wire-harnessing success rates:} The robot harnesses two wires to solve task 1 and task 2 in a sequence. The robot executes task 2 only if the task 1 is successfully solved.}
  \label{table:multi-tasks}
\end{table}

Lastly, we evaluated our method in a challenging multi-wire harnessing scenario, wherein the robot had to harness two wires into desired shapes for tasks 1 and 2. The snapshots of the harnessing sequence are shown in Fig.~\ref{fig:snapshots}. As Table~\ref{table:multi-tasks} illustrates, our method replicated the success of the single-wire task in the first harnessing attempt. However, the second task's complexity increased due to the presence of an already inserted wire, which raises the likelihood of entanglement. Despite this, our method successfully completed 8 out of 9 trials for the second task, showcasing substantial robustness. In contrast, the No-twist baseline could not advance past the first task, and the Linear model baseline, struggling with tension regulation, achieved only 6 successful multi-wire harnessing out of 10 attempts.
\section{Conclusion}

In this paper, we introduced a robust single-robot wire-harnessing pipeline, diverging from previous approaches that relied on dual-robot arms or tactile sensing. Our method employs the robot's twisting motion to generate the tension necessary for inserting the wire into clamps, utilizing only the built-in F/T sensor. The approach comprises four key components: 1) a Koopman operator-based MPC for wire following while maintaining the desired tension, 2) a harnessing motion planner to determine a sequence of waypoints, 3) a set of insertion primitives for securing the wire in clamps, and 4) a fix-point switching module to update the wire's constraints. We assessed our method using an industrial-level wire harnessing testboard, where it demonstrated robust performance and high success rates, and outperforms other baselines. The capability to handle multi-wire harnessing tasks further underscores the effectiveness of our method.

However, our method is not without limitations: 1) It assumes the wire is pre-inserted and an initial wire pose that is distanced from the clamps to avoid tangling with the task board. 2) The learned wire dynamics model is learned on a specific wire type and may not generalize directly to wires of different sizes and materials. 3) The continuous twisting and stretching motions could lead to material fatigue and damage. Future work will explore learning initial sorting and insertion motions through reinforcement or imitation learning to set the stage for the harnessing task. Additionally, we aim to develop online adaptation techniques to update the wire dynamics model in real-time \cite{zhang2023efficient,sun2021online,wang2022safe,wang2022offline}, enabling the system to adjust to new wires during operation. Also, the desired tension force can be adjusted during execution to minimize the potential wire damage, similar to \cite{zhang2024bridging}.

\addtolength{\textheight}{-12cm}   




\bibliographystyle{IEEEtran}
\bibliography{IEEEexample}

\end{document}